%% file: main.tex
\documentclass[10pt,twocolumn,letterpaper]{article}

\usepackage{cvpr}
\usepackage{times}
\usepackage{epsfig}
\usepackage{graphicx}
\usepackage{amsmath}
\usepackage{amssymb}
\usepackage{multirow}

\usepackage[pagebackref=true,breaklinks=true,letterpaper=true,colorlinks,bookmarks=false]{hyperref}

\def\cvprPaperID{1114} 

\ifcvprfinal\fi

\cvprfinalcopy 

\def\cvprPaperID{1114} 

\ifcvprfinal\fi

\title{A Pose-Sensitive Embedding for Person Re-Identification\linebreak with Expanded Cross Neighborhood Re-Ranking}

\author{M. Saquib Sarfraz$^1$, Arne Schumann$^{2,1}$, Andreas Eberle$^1$, Rainer Stiefelhagen$^1$\\
$^1$Karlsruhe Institute of Technology, $^2$Fraunhofer IOSB\\
Karlsruhe, Germany\\
{\tt\small  \{firstname.lastname\}@kit.edu}}

\begin{document}
\pagenumbering{gobble}
\maketitle

\input{sections/abstract}
\input{sections/introduction}
\input{sections/related_work}
\input{sections/methodology}

\input{sections/evaluation}

\input{sections/conclusion}

{\small
\bibliographystyle{ieee}
\bibliography{egbib}
}

\end{document}

%% file: sections/abstract.tex
\begin{abstract}
Person re-identification is a challenging retrieval task that requires matching a person's acquired image across non-overlapping camera views. In this paper we propose an effective approach that incorporates both the fine and coarse pose information of the person to learn a discriminative embedding. In contrast to the recent direction of explicitly modeling body parts or correcting for misalignment based on these, we show that a rather straightforward inclusion of acquired camera view and/or the detected joint locations into a convolutional neural network helps to learn a very effective representation. To increase retrieval performance, re-ranking techniques based on computed distances have recently gained much attention. We propose a new unsupervised and automatic re-ranking framework that achieves state-of-the-art re-ranking performance. We show that in contrast to the current state-of-the-art re-ranking methods our approach does not require to compute new rank lists for each image pair (\eg., based on reciprocal neighbors) and performs well by using simple direct rank list based comparison or even by just using the already computed euclidean distances between the images. We show that both our learned representation and our re-ranking method achieve state-of-the-art performance on a number of challenging surveillance image and video datasets. 
Code is available at \url{https://github.com/pse-ecn}.
\vspace{-.5cm}
\end{abstract}

%% file: sections/introduction.tex
\section{Introduction}
\label{sec:intro}
\begin{figure}
\centering
\includegraphics[width=\columnwidth]{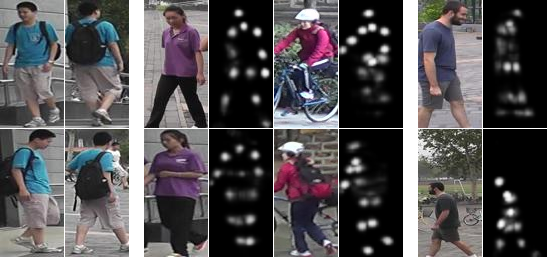}
\caption{Camera view and body pose can significantly alter the visual appearance of a person. A different view might show different aspects of the clothing (\eg a backpack) while an altered pose may lead to body parts (\eg arms or legs) being located at different positions of the image. Pose information can also help guide the attention of a re-id approach in case of mis-aligned detections.}
\label{fig:intro}
\vspace{-.5cm}
\end{figure}

Person re-identification (re-id) in non-overlapping camera views poses a difficult matching problem. Most previous solutions try to learn the global appearance of a persons using Convolutional Neural Networks (CNNs) by either applying a straightforward classification loss or using a metric learning loss. To better learn local statistics, the same has been applied to image regions, \eg by using horizontal stripes or grids \cite{li2014deepreid, cheng2016person}. Because of the inherent challenge of matching between different views and poses of a person, there is no implicit correspondence between the local regions of the images (see Figure \ref{fig:intro}). This correspondence can be established by explicitly using full body pose information for alignment \cite{su2017pose} or locally through matching corresponding detected body parts \cite{zhao2017spindle,zhao2017deeply}. Using this local or global person description by incorporating the body pose or body parts information can strongly benefit person re-id.

In this work we show that incorporating a simple cue of the person's coarse pose (\ie the captured view with respect to the camera) or the fine body pose (\ie joint locations) suffice to learn a very discriminative representation with a simple classification loss. We present an appealing design choice to incorporate these cues and show its benefit in the performance gain over state-of-the-art methods on large and challenging surveillance benchmarks. We demonstrate that learning and combining view specific feature maps on a standard underlying CNN architecture results in a significantly better re-id embedding. Similarly an incorporation of body joint locations as additional input channels helps to increase the re-id accuracy.

For improving person retrieval, after computing the initial distances, a re-ranking step can often improve ranking quality by a good margin. Re-ranking has seen a renewed interest in recent years \cite{ma2014query,garcia2015person,leng2015person,ye2016person,zhong2017re}. The re-ranking problem is formulated as re-estimating the distances between probe and gallery images such that more correct results are ranked at the top of the returned lists. In recent proposals this was generally achieved by exploiting the similarity of the lists of top k nearest neighbors of both the probe and gallery image in question. Among the state-of-the-art re-ranking methods these neighborhood lists are often recomputed for each image pair, based on the common or reciprocal neighbors \cite{ye2015coupled,bai2016sparse,zhong2017re}. This makes it more computationally demanding to recompute the distances between these varying length lists.

A second contribution of this work is a new re-ranking method that introduces the concept of expanded cross neighborhood distance. The method aggregates the distances of close neighbors of the probe and the gallery image, where the distance can simply be the direct euclidean distance or the distance based on the rank lists. We show that within this more general framework of re-ranking simple rank list comparison based on the directly obtained rank lists achieves state-of-the-art re-ranking performance without the requirement to recompute new rank lists.

In summary, our contributions are threefold: 1) We propose a new CNN embedding which incorporates coarse and fine-grained person pose information.  2) We propose a new unsupervised and automatic re-ranking method that achieves larger re-ranking improvements than previous methods.
3) Our pose-sensitive person re-id model and our re-ranking method set a new state-of-the-art on four challenging datasets. We also demonstrate the scalability of our approach with very large gallery sizes
and its performance for person search in full camera images.

%% file: sections/related_work.tex
\section{Related Work}
\label{sec:relwork}

In recent years many state-of-the-art re-id results have been achieved by approaches relying on feature embeddings learned through CNNs \cite{zhao2017spindle,hermans2017defense,zhang2017deep,JLML}. We focus our discussion of related approaches to those which include a degree of pose information, as well as re-ranking methods.

\vspace{.1cm}
\noindent\textbf{Re-Id using Pose} 
A person's body pose is an important cue for successful re-identification. The popular SDALF feature by Farenza \etal~\cite{farenzena2010person} uses two axes dependent on the body's pose to derive a feature description with pose invariance. Cho \etal~\cite{cho2016improving} define four view angles (front, left, right, back) and learn corresponding matching weights to emphasize matching of same-view person images.
A more fine-grained pose representation based on Pictorial Structures was first used in \cite{cheng2011custom} to focus on matching between individual body parts.
More recently, the success of deep learning architectures in the context of re-id has lead to several works that include pose information into a CNN-based matching. 
In \cite{zheng2017pose} Zheng \etal propose to use a CNN-based external pose estimator to normalize person images based on their pose. The original and normalized images are then used to train a single deep re-id embedding.
A similar approach is described by Su \etal in \cite{su2017pose}. Here, a sub-network first estimates a pose map which is then used to crop the localized body parts. A local and a global person representation are then learned and fused.
Pose variation has also been addressed by explicitly detecting body parts through detection frameworks \cite{zhao2017spindle}, by relying on visual attention maps \cite{rahimpour2017person}, or body part specific attention modeling \cite{zhao2017deeply}.

In contrast to our proposed method, these works mostly rely only on fine-grained pose information. Furthermore, these approaches either include pose information by explicitly normalizing their input images or by explicitly modeling part localization and matching these in their architecture. In contrast to this, our approach relies on confidence maps generated by a pose estimator which are added as additional channels to the input
image. This allows a maximum degree of flexibility in the learning process of our CNN and leaves it to the network to learn which body parts are relevant and reliable for re-id. Apart from this fine-grained pose information we show that a more coarse pose cue turns out to be even more important and can be effectively used to improve the re-id performance.

\begin{figure*}
\centering
\includegraphics[width=\textwidth]{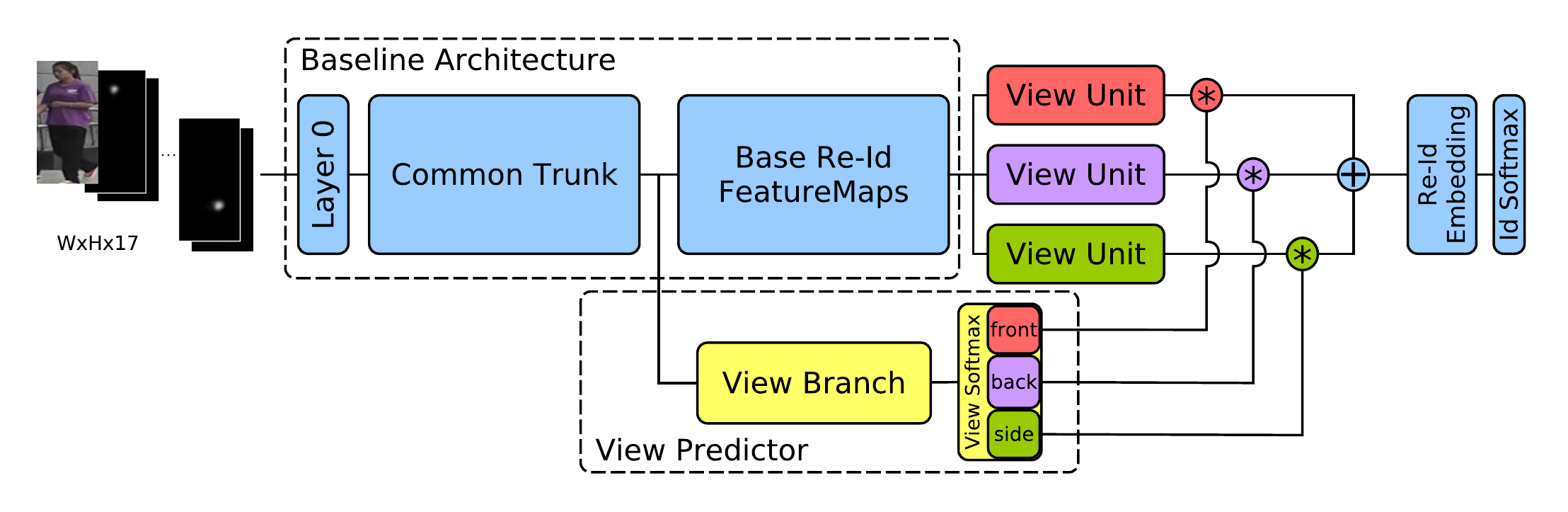}
\caption{Overview of our pose sensitive embedding (PSE) architecture. As baseline architecture we employ either ResNet-50 or Inception-v4. Pose information is included through detailed body joint maps in the input, as well as through a coarse view predictor.}
\label{fig:overview}
\vspace{-.4cm}
\end{figure*}

\noindent\textbf{Re-Ranking}
In the recent years, re-ranking techniques are drawing more and more attention in the area of person re-id.
Shen \etal~\cite{shen2012object} use k-nearest neighbors (k-NN) to produce new rank lists and recompute distances based on these. Garcia \etal~\cite{garcia2015person} propose to jointly learn the context and content information in the ranking list to remove candidates in the top neighbors and improve performance of person re-id. \cite{leng2015person} extends this to revise the initial ranking list with a new similarity obtained from fusion of content and contextual similarity.
Li \etal~\cite{li2012common} first proposed to use the relative information of common nearest neighbors of each image to improve re-ranking. Ye \etal~\cite{ye2015coupled} combined the common nearest neighbors of global and local features as new queries and revise the initial ranking list by aggregating these into new ranking lists. Using the similarity and dissimilarity cues from the neighbors of different baseline methods \cite{ye2016person} proposes a ranking aggregation algorithm to improve person re-id.
In contrast to common neighbors, Jegou \etal~\cite{jegou2007contextual}  use reciprocal neighbors (\ie common neighbors that reciprocate in a k-neighborhood sense) and propose to compute a contextual dissimilarity measure (CDM). \cite{qin2011hello} formally uses the k-reciprocal neighbors to compute ranking lists. Most recent state-of-the-art re-ranking methods are based on computing these rank list comparisons using a generalized Jaccard distance.
To overcome the associated complexity of computing intersection and unions of underlying variable length lists, Sparse Contextual Activation (SCA) \cite{bai2016sparse} encodes the neighborhood set into a sparse vector and then computes the distance. To reduce the false positives and noise in the original ranked lists, more context is included by forming new rank lists based on reciprocal neighbors \cite{jegou2007contextual}\cite{qin2011hello}\cite{zhong2017re}. Zhong et al \cite{zhong2017re} use the k-reciprocal lists and compute the Jaccard distance by using SCA encoding. They then propose to fuse this distance with the original distance to obtain the final re ranking. Note, while reciprocal list based comparisons provides the current best re-ranking scores, it requires an additional complexity of recomputing the reciprocal rank lists for each image pair.

In contrast to common or reciprocal neighbors and producing new rank lists based on these, we propose the concept of expanded neighbors and aggregating their cross distances among the images in a pair. We show that this results in a more effective re-ranking framework while not requiring to re-compute new rank-lists for each image pair.

%% file: sections/methodology.tex
\section{Pose-Sensitive Embedding}
\label{sec:embedding}
A person's pose and orientation to the camera can greatly affect the visual appearance in the image.
Explicitly including this information into the learning process of a re-id model can thus
often increase the resulting accuracy. Previous works have relied on either fine-grained pose 
information (\eg joint keypoints) or coarse information (\eg orientation to the camera). In this 
section we describe two new methods for including both levels of granularity into a pose-sensitive
embedding. Both methods can be simultaneously incorporated into the same baseline CNN architecture and our experiments show that a combination of the two achieves a higher accuracy than 
either one alone. An overview of our CNN architecture with both types of pose information is 
depicted in Figure \ref{fig:overview}.

\subsection{View Information}
\label{sec:view}

We use the quantization [`front', `back', `side'] of a person's orientation to the camera
as coarse pose information. Since this information is dependent on the camera, as well as the person,
we call it \emph{view information} in the remainder of this work.

Our inclusion of view information into
the re-id embedding is based on our prior work \cite{BMVC} on semantic attribute recognition.
A ternary view classifier is added as a side-branch of our main re-id CNN. The tail part of the main CNN is then split into three equivalent units by 
replicating existing layers. The view classifier's three view prediction scores are used to weight
the output of each of these units. This modulates the gradient flowing through the units, \eg for a 
training sample with a strong `front' prediction, mainly the unit weighted by the front-weight will 
contribute to the final embedding and thus only this unit will receive a strong
gradient update for the current training sample. This procedure allows each unit to learn a feature map
specialized for one of the three views. Importantly, and in contrast to \cite{BMVC}, we do not weight 
and fuse final embeddings or prediction vectors but apply the weights to full feature maps which are then combined into the final embedding. This achieves a more robust representation.

We cannot generally assume to have view annotations available on the re-id 
dataset we want to train our embedding on. Thus, we pretrain a corresponding view classifier on the separate RAP \cite{RAP} pedestrian dataset which provides such annotations. We then directly transfer the classifier to our re-id model. Low-level features (\ie early layers) can be shared between the view predictor and the re-id network in order to reduce model complexity.

In our default ResNet-50 architecture the view predictor branch is split off from the main network after the third dimensionality reduction step (\ie at feature map dimensions $28\times28\times256$). We then apply three consecutive convolutions with step sizes 2, 2, and 5 to reduce the dimension further (to $1\times1\times1024$). The resulting feature vector is used to predict the view using a three-way softmax. As view units we use three replications of the ResNet Block-4. The $7\times7\times2048$ dimensional fused output of the units is pooled and fed to a fully connected layer which yields our 1536 dimensional embedding.

\subsection{Full Body Pose}
\label{sec:pose}

As fine-grained representation of a person's pose we use the locations of 14 main body
joint keypoints. To obtain this information we use the off-the-shelf DeeperCut \cite{deepercut}
model. In contrast to prior use of pose information for re-id, we do not use 
this information to explicitly normalize the input images. Instead, we include the information into the 
training process by adding an additional input channel for each of the 14 keypoints. These channels serve to guide the CNN's attention and help learn how to best apply the body joint information into the resulting embedding. To further increase this flexibility, we do not rely on the final keypoint decisions
of the DeeperCut approach, but instead provide our re-id CNN with the full confidence map for
each keypoint. This prevents any erroneous input based on hard keypoint decisions and leaves our model
the chance to compensate for, or at least recognize, unreliable pose information.

\subsection{Training Details}
We initialize all our CNNs with weights pretrained for ImageNet classification.
In order to train a model with view information (Section \ref{sec:view}) we start by fine-tuning only the view-predictor branch on the RAP dataset \cite{RAP}. Next we 
train only the view units and the final person identity classification layer on the target 
re-id dataset. The 
weights of the view predictor and all layers before the view units are fixed for this stage. This 
allows the randomly initialized view units and final layers to adapt to the existing weights of
earlier layers.

When training an embedding including full body pose information (Section \ref{sec:pose}) the ImageNet weights do not match the size of our input, due to the additional 14 keypoint channels. To adapt the network for 17 channel inputs we
start our training by fine-tuning only the first layer (Layer 0 in Figure \ref{fig:overview}), and the final person identity classification
layer which are both initialized randomly. The remainder of the network remains fixed. Once these
two layers are adapted to the rest of the network (\ie convergence is observed), we proceed by fine-tuning the entire network.

For our final pose sensitive embedding (PSE) we combine both types of pose information into one network as depicted in Figure \ref{fig:overview}.
We initialize our training with the full body pose model described in the previous section and add the view predictor onto it. The view predictor is fine-tuned on the RAP dataset with pose maps and can benefit from the additional full body pose information. Further fine-tuning of the re-id elements of the network is then performed on the target re-id dataset as described above.

For all our CNN embeddings we employ the same training protocol. Input images are normalized to channel-wise zero-mean and a standard variation of 1. Data augmentation is performed by resizing images to 105\% width and 110\% height and randomly cropping the training sample, as well as random horizontal flip (this is the main reason why we do not differentiate between left and right side views). Training is performed using the Adam optimizer at recommended parameters with an initial learning rate of 0.0001 and a decay of 0.96 every epoch.

\section{Expanded Cross Neighborhood Distance based Re-Ranking}
\label{sec:ECN}
In this section we introduce the concept of Expanded Cross Neighborhood (ECN) distance which can provide a very high boost in performance while not strictly requiring rank list comparisons. We show that, for an image pair, accumulating the distances of only the immediate two-level neighbors of each image with the other image results in a promising re-ranking. Within this cross neighborhood based distance framework, the underlying accumulated distances can be just the original euclidean distances or the re-calculated rank-list based distances. We also show that within this framework using a simple list comparison measure on the initially obtained rank lists achieves the state of the art re-ranking performance. Our proposal is fully automatic, unsupervised and can work well without requiring to compute new rank lists.

Formally, given a probe image $p$ and a gallery set $G$ with $B$ images $G=\left \{ g_i \mid i=1,2, \cdots , B \right \}$, the euclidean distance between $p$ and each of the gallery $g_i$ is $\lVert \mathbf{p}-\mathbf{g_i} \rVert_{2}^{2}$. Computing pairwise distance between all images in the gallery and probe sets, the initial ranking $\mathcal{L}(p,G)=\left \{ g_1^o, \cdots , g_B^o \right \} $ for each image is then obtained by sorting this distance in an increasing order. 

Given such initial rank lists $\mathcal{L}$ of all the images in the gallery and probe sets, we define the expanded neighbors of the probe $p$ as the multiset $N(p,M)$ such that:
\begin{equation}\label{eq:ecn1}
N(p,M) \leftarrow \{N(p,t), N(t,q)\}
\end{equation}
where $N(p,t)$ are the top $t$ immediate neighbors of probe $p$ and $N(t,q)$ contains the top $q$ neighbors of each of the elements in set $N(p,t)$:
\begin{equation}\label{eq:ecn2}
\begin{aligned}
 N(p,t)&=\{g_i^o \mid i=1,2, \cdots, t \} \\
 N(t,q)&=\{ N(g_i^o,q), \cdots, N(g_t^o,q)\} 
\end{aligned}
\end{equation}

A similar expanded neighbors multiset can be obtained for each of the gallery images $N(g_i,M)$ in terms of its immediate neighbors and their neighbors. The total number of neighbors $M$ in the set $N(p,M)$ or $N(g_i,M)$ is $M= t + t \times q$. Finally the Expanded Cross Neighborhood (ECN) distance of an image pair $(p,g_i)$ is defined as
\begin{equation}\label{eq:ecn3}
ECN(p,g_i)=\frac{1}{2M}\sum_{j=1}^{M} d(pN_j,g_i) + d(g_iN_j,p)
\end{equation}

where $pN_j$ is the $jth$ neighbor in the probe expanded neighbor set $N(p,M)$ and $g_iN_j$ is the $jth$ neighbor in the $ith$ gallery image expanded neighbor set $N(g_i,M)$. The term $d(\cdot)$ is the distance between that pair. One can see that the ECN distance, above, just aggregates the distances of the expanded neighbors of each of the image in pair with the other. While we show in our evaluation that using the direct euclidean distance in Equation \ref{eq:ecn3} results in a similar improvement in the rank accuracies, one can also use a more robust rank-list based distance to further enhance the performance, especially in terms of the mean average precision (mAP). These distances can be computed directly from the initial paired distance matrix or the resulting initial rank lists. Recent re-ranking proposals use the Jaccard distance for the list comparison which is computationally expensive, here we propose to use a rather simple list comparison similarity measure proposed by Jarvis and Patrick \cite{jarvis1973clustering}, and also successfully employed in a face verification task in \cite{schroff2011pose}. The list similarity is measured in terms of the position of top \textit{K} neighbors of the two lists. For a rank list with $B$ images, let $pos_i(b)$ denote the position of image $b$ in the ordered rank list $\mathcal{L}_i$. In terms of considering only the first $K$ neighbors in the list, the Rank-list similarity $R$ is given by:

\vspace{-.5cm}
\begin{equation}\label{eq:ecn4}
R(\mathcal{L}_i,\mathcal{L}_j)=\sum_{b=1}^{B} \left [ K+1-pos_i(b) \right ]_+ \times \left [ K+1-pos_j(b) \right ]_+
\end{equation}

Here, $\left [\cdot \right ]_+=max(\cdot,0)$. This measure ensures to base similarity in terms of top $K$ neighbors while taking into account their position in the list. From an implementation point, this rank list similarity can effectively be computed from the initially obtained rank lists by single matrix multiplication and addition operations. To use this in Equation \ref{eq:ecn3}, we convert it into the distance $d = 1 - R^*$ where  $R^*$ denotes the minmax scaling of values in $R$. Finally the parameters $t$, $q$ and $K$ (in case of using the rank-list distances) for computing the final ECN distance are set to $t=3$, $q=8$ and $K=25$. while we show that these parameters choices are very stable in terms of performance on a number of different sized datasets, one can intuitively also see that using the strongest top neighbors in the first level ($t$) and expanding these to few more at the second level ($q$) makes sense. Note since our neighbors' of neighbor expansion only looks for the first and second level of neighbors, we do not need to compute an expensive KD-tree or neighborhood graphs to get these expanded sets in Equation \ref{eq:ecn1}, we can readily obtain these from the initially computed ordered rank list matrix.

%% file: sections/evaluation.tex
\newcommand{\f}[1]{\textbf{#1}} 
\newcommand{\np}[0]{$^\dag$}    
\newcommand{\tnote}[1]{\tiny{\textcolor{red}{#1}}}

\section{Evaluation}
\label{sec:eval}
We report results using the standard cross camera evaluation in the single-query setting. Accuracy is measured by rank scores, obtained from cumulated matching characteristics (CMC), and mean average precision (mAP).

\noindent{\textbf{Datasets:}} We evaluate our approach on four datasets, Market-1501 \cite{market} (Market), Duke-MTMC-reID \cite{Duke} (Duke), MARS \cite{MARS} and PRW \cite{{PRW}}.

The Market-1501 (Market) dataset consists of 32,668 bounding boxes of 1,501 distinct persons generated by a person detector on videos from six cameras. 751 persons are used for training and 750 for testing. The training set contains 12,936 images, the gallery set 19,732 images, and the query set has 3,368 images.

The Duke-MTMC-reID (Duke) dataset is created from data of eight cameras. Of 1,812 people in the data 1,404 occur in more than one camera. Training and test sets both consist of 702 persons. The training set includes 16,522 images, the gallery 17,661 images, and the query set 2,228 image. Person bounding boxes in the Duke dataset are manually annotated.

The MARS dataset consists of 20,478 tracklets of 1,261 re-occurring persons. Including 3,248 distractor tracklets this brings the total number of person images in the dataset to 1,191,993 with a train/test split of 509,914/681,089 images of 625 and 636 persons, respectively. This dataset is well suited to evaluate the performance of a re-id approach for person track retrieval.

The PRW dataset consists of 11,816 frames of video data.
The images are annotated with 43,110 person bounding boxes of which 34,304 are assigned one of 932 person IDs. For training 5,134 frames including 482 different persons are available. At test time 2,057 cropped query images of persons must be found in a gallery of 6,112 full images. The PRW dataset allows for an evaluation of the robustness of a re-id method to false positive or mis-aligned person detections.

In order to compare to related approaches we split our evaluation into three parts. In Sections \ref{sec:eval-pse} and \ref{sec:eval-rerank} we investigate key components of our pose-sensitive embedding and re-ranking, respectively. In Section \ref{sec:eval-sota} we compare our proposed embedding and re-ranking with state-of-the-art approaches. We also demonstrate the robust performance of our approach against detector errors and its scalability for very large galleries.

\begin{table*}[htb]
\centering
\begin{tabular}{ | c | c | ccccc | ccccc |} 
    \hline
	CNN & Method & \multicolumn{5}{|c|}{Market-1501} & \multicolumn{5}{|c|}{Duke}  \\
	                           & &      mAP & R-1 & R-5 & R-10 & R-50  &    mAP & R-1 & R-5 & R-10 & R-50 \\
	\hline\hline
    Inception-v4 & Baseline   &     51.9 &     75.9 &  89.8	& 92.5 & 97.3 &    36.6 &     61.8 & 74.8&	79.8	& 89.4  \\
	             & Views only &     61.9 &     81.5  &  92.3	& 94.9	& 98.1 &     40.3 &     62.7   & 76.6&	81.1	& 90.3 \\
		         & Pose only  &     60.9 &     81.7  &  91.8	&94.4 &	97.9  &     48.2 &     70.5  & 81.9 &	86.1&	92.7   \\
                 & PSE        & \f{64.9} & \f{84.4}   & \f{93.1}	&\f{95.2} &	\f{98.4} & \f{50.4} & \f{71.7}   & \f{83.5}	& \f{87.1} &	\f{93.1} \\
	\hline
	ResNet-50    & Baseline   &     59.8 &     82.6 &  92.4	& 94.9 &	98.2  &       50.3 &     71.5   &83.1&	87.0	& 94.1  \\
           		 & Views only &     66.9 & \f{88.2} &   \f{ 95.4}	& \f{97.2}	& 98.9  &     56.7 &     76.9  &  87.3&	90.7	& 95.7    \\
		   		 & Pose only  &     61.6 &     82.8 &   93.1	& 95.5	& 98.3  &      53.1 &     73.4  &  84.5	&88.1 &	94.3  \\
           		 & PSE		  & \f{69.0} &     87.7 &  94.5	& 96.8 &	\f{99.0}  &    \f{62.0} & \f{79.8}  &  \f{89.7} & \f{92.2} & \f{96.3}\\
	\hline
\end{tabular}
\vspace{.1cm}
\caption{Comparison of different types of pose information. While views and full body pose individually lead to notable improvements, a combination of both often results in further improvements.}
\label{tab:ablation_pose}
\vspace{-.2cm}
\end{table*}
\subsection{Study of Pose Information}
\label{sec:eval-pse}
We investigate the usefulness of including different granularities of
pose information into the CNN by performing separate experiments with only view information, 
only pose information, and a combination of both. Experiments are performed on Market and Duke. To show that our proposal is not strictly dependent on the underlying CNN architecture, besides using our main ResNet-50 base CNN, we also show results on the popular Inception-v4 CNN. For Inception-v4, the view predictor is branched out at the earlier Reduction-A block and view units are similarly added by using three Inception-C blocks at the end. Results of
our experiments are given in Table \ref{tab:ablation_pose}.

Compared to a baseline without any explicitly modeled pose information, inclusion of either views or pose significantly increases the accuracy of the resulting feature embedding. This observation holds across both datasets, as well as both network architectures. For the ResNet model, the view information results in a larger absolute improvement of about 6-7\% in mAP on both datasets while the pose information leads
only to an improvement of about 2-3\% in mAP. Results for the Inception-v4 model are less consistent.
Both types of information still achieve large improvements but on the Market dataset the absolute 
improvement for both types lies around 10\% in mAP while on Duke the 11\% mAP improvement through pose 
information clearly outperforms the 4\% gained by including view information. 

Finally, a combination of the two types of information leads to a further consistent increase in mAP compared to the best result of either individual pose information. For instance, on the base ResNet-50 model, the combination achieves a further improvement in mAP of 2.1\% and 5.3\% on Market and Duke, respectively. Similarly, on the base Inception-v4 model the combination further improves the mAP by 3\% on Market and 2.2\% on Duke. This clearly indicates that our methods of including different degrees
of pose information complement each other.

\begin{figure}[t]
\centering
\includegraphics[scale=1.5]{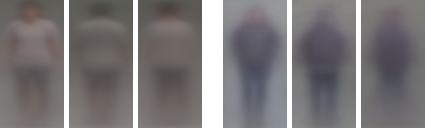}
\caption{Mean images of Market-1501 (left) and Duke (right) test sets using predictions of the PSE model's view predictor. The images show front, back and side view from left to right.}
\label{fig:means}
\vspace{-.5cm}
\end{figure}

\noindent{\textbf{View-predictor performance:}} The performance of the trained ResNet-50 view predictor on the annotated test set of RAP dataset is 82.2\%, 86.9\% and 81.9\% on front, back, and side views, respectively. In order to illustrate its performance on our target re-id dataset we display mean images in Figure \ref{fig:means}. These are obtained by averaging all images, on the test set of the target dataset, which are classified as front, left, or side. This visualization gives an impression of the view prediction accuracy on the target re-id data in the absence of annotated view labels. In the frontal mean image a skin-colored face region is clearly discernible, indicating that the majority of images were in fact frontal ones. Similarly, the back mean image correctly shows the backside of a person. The side view is more ambiguous, aside from the possible view predictor errors, mainly because we group left and right side into one combined class.

\subsection{Study of Re-Ranking}
\label{sec:eval-rerank}

In Table \ref{tab:ablation_reranking} we compare several configurations of our proposed ECN 
re-ranking with other popular re-ranking methods across the Market, CUHK03 (detected) \cite{li2014deepreid} and MARS datasets. Note that the CUHK03 includes both the labeled and detected (using a person detector) person bounding boxes. We chose the CUHK03 (detected) as it is more challenging. We evaluate CHUK03 under the new fixed train/test protocol as used in \cite{zhong2017re} \cite{yu2017divide}.
To compare with the published results of several re-ranking methods on these datasets, we use the same baseline features, 2,048-dim ID-discriminative embedding provided by \cite{zhong2017re}. We compare with the previous re-ranking techniques for object retrieval and person re-id including contextual dissimilarity measure (CDM) \cite{jegou2007contextual}, spatially constrained (k-NN) re-ranking \cite{shen2012object}, Average query expansion (AQE) \cite{chum2007total}  and the current state-of-the-art Sparse Contextual Activation (SCA) \cite{bai2016sparse} , k-reciprocal encoding (k-reciprocal) \cite{zhong2017re} and its direct multiplicative application Divide and Fuse (DaF) \cite{yu2017divide}.
As shown our ECN re-ranking achieves a consistent improvement in performance across all three datasets on both mAP and rank-1 metrics. 

We provide the performance of the different components of our ECN framework. As shown in Table \ref{tab:ablation_reranking}, only using the rank-list distance of Equation \ref{eq:ecn4} (rank-dist) still provides meaningful performance gains. Within the ECN framework just using the direct euclidean distances in Equation \ref{eq:ecn3} `ECN (orig-dist)' results in similar high performance gains in the rank-1 scores, in fact better than the state-of-the-art k-reciprocal \cite{zhong2017re} method that uses the reciprocal list comparisons with local query expansion and fusion of rank and euclidean distances. As this does not involve computing any rank list based comparison, this result is an additional very attractive outcome of our proposal. Finally our ECN re-ranking using the simple rank-list comparison of Equation \ref{eq:ecn4} as distance in the ECN Equation \ref{eq:ecn3} provides the best results and improves the mAP further.

\newcommand{\mcl}[1]{\multicolumn{2}{|l|}{#1}}
\begin{table}[t]
\centering
\resizebox{\columnwidth}{!}{
\begin{tabular}{ | l|l | cc | cc | cc |}
    \hline
    \mcl{Re-Ranking} & \multicolumn{2}{|c|}{Market-1501} & \multicolumn{2}{|c|}{CUHK03} & \multicolumn{2}{|c|}{MARS} \\
    \mcl{}           & \multicolumn{2}{|c|}{IDE-R}       & \multicolumn{2}{|c|}{IDE-R}  & \multicolumn{2}{|c|}{IDE-C} \\
    \mcl{}       					    &    mAP &    R-1 &    mAP &    R-1 &    mAP &    R-1 \\
	\hline\hline
    \mcl{None}        					&   55.0 &   78.9 &   19.7 &   21.3 & 	41.2 &   61.7 \\
    \mcl{AQE \cite{chum2007total}}     	&      - &      - &      - &      - & 	47.0 &   61.8 \\
    \mcl{CDM \cite{jegou2007contextual}}&   56.7 &   79.8 &   20.6 &   22.9 & 	44.2 &   62.1 \\
    \mcl{K-NN \cite{shen2012object}}    &   60.3 &   79.5 &   22.9 &   24.3 & 	   - & 		- \\
    \mcl{SCA  \cite{bai2016sparse}}    	&   68.9 &   79.8 &   26.6 &   24.7 &      - & 		- \\
    \mcl{k-reciprocal \cite{zhong2017re}}&  70.4 &   81.4 &   27.3 &   24.9 &   51.5 &   62.8 \\
    \mcl{DaF \cite{yu2017divide}}  		&\f{72.4}&   82.3 &   30.0 &   26.4 &      - & 	    - \\
    \hline
    \multirow{3}{*}{\rotatebox{90}{\hspace{-.1cm}Our}}
    & Rank dist (Eq. \ref{eq:ecn4})     &   66.1 &   80.3 &   25.0 &   25.3 &   48.7 &   62.2 \\
    & ECN (orig-dist)				    &   66.7 &   81.7 &   27.5 &   25.9 &   50.1 &   64.7 \\
    & \f{ECN (rank-dist)} 				&   71.1 &\f{82.3}&\f{30.2}&\f{27.3}&\f{53.2}&\f{64.6}\\
	\hline
\end{tabular}
}
\vspace{.1cm}
\caption{Comparison of the proposed ECN re-ranking method with state-of-the-art on three datasets, Market-1501, CHUK03 (detected) and MARS. Baseline features: 2,048-dim ID-discriminative Embedding fine tuned on Resnet (IDE-R) and CaffeNet (IDE-C) \cite{zhong2017re}.}
\label{tab:ablation_reranking}
\vspace{-.4cm}
\end{table}

\noindent\textbf{Parameters impact:} In all of our evaluations presented in Table \ref{tab:ablation_reranking} as well as in Table \ref{tab:SOTA}, the ECN parameters are set to $t$=$3$ and $q$=$8$.  Given the very different number of images in query and test sets of the used datasets, the results show the stability of these parameters. We studied the impact of changing these on Market and Duke datasets and found that it is subtle in the range for $t \in [2,4]$ and $q\in [4,10]$, the performance drops between ${\sim} 0.2$-$0.8\%$ on different combinations within this range. Similarly the impact of parameter $K$ in Equation \ref{eq:ecn4} works well within $K\in [10,30]$, with better performance when $K>20$ on all three large datasets Market, MARS and Duke. The jitter in accuracies with changing K in this range stays within ${\sim}{\pm}2\%$. 

Since CUHK03 is a relatively small dataset, both DaF \cite{yu2017divide} and k-reciprocal \cite{zhong2017re} report results on CUHK03 by using different parameters values for their methods than used for the other datasets. While we used the same ECN parameters of $t$=$3$, and $q$=$8$ on CUHK03, we obtained higher performance with the parameter $K$=$10$ instead of $K$=$25$ (as used on all other datasets) for the rank-list distance in Equation \ref{eq:ecn4}. The reported results in table \ref{tab:ablation_reranking} on CHUK03 dataset are with $K$=$10$, however with $K$=$25$, we still get better performance than the most state-of-the-art methods with mAP of $28.4\%$ and rank-1 of $26.0\%$.  

\noindent\textbf{Complexity analysis:} The computational complexity of ECN is $\mathcal{O}(N^2logN)$ (same as other re-ranking methods) but it executes fewer computation steps by avoiding re-computing the neighbors' lists for each image pair. In its variant with ECN (orig-dist) it offers close improvements without having to re-compute the rank lists based distance (hence even fewer steps). For example, on the large Duke dataset (re-ranking on 19,889 total images), computation times averaged over five runs are $124.6$s for the related work k-reciprocal \cite{zhong2017re} while $115.3$s and $73.2$s for our ECN (rank-dist) and ECN (orig-dist) respectively.

\subsection{State-of-the-art}
\label{sec:eval-sota}





In Table \ref{tab:SOTA} we compare the performance of our approach with the published state-of-the-art on the three popular datasets (Market, Duke, and MARS). In the top section of the table we compare approaches without any re-ranking to our pose-sensitive embedding. The embedding achieves top accuracy on both MARS and Duke datasets. On the Market dataset our embedding performs slightly worse than the DPFL \cite{DPFL} approach which employs two or more multi-scale embeddings. Across all three datasets the increase in mAP achieved by including pose information on the base ResNet 
ranges from 7.4\% to 11.7\%. In the bottom section of Table \ref{tab:SOTA} we include the best published methods with re-ranking. In combination with our proposed re-ranking scheme we set a new state-of-the-art on all three datasets by large margins. On Market we increase mAP by 11.4\%, on Duke by 19.2\%, and on MARS by 4.5\%.

\begin{table*}
\centering
\resizebox{.8\textwidth}{!}{ 
\begin{tabular}{ | l|l|l | cc | cc | cc |}
    \hline
    \multicolumn{3}{|c|}{Method} & 
	\multicolumn{2}{|c|}{Market-1501} & 
    \multicolumn{2}{|c|}{Duke}  &
    \multicolumn{2}{|c|}{MARS}  \\

    \multicolumn{3}{|c|}{}                  & mAP  & R-1  & mAP  & R-1  & mAP  & R-1  \\
	\hline\hline
    \mcl{P2S\cite{P2S}}                     & CVPR17        & 44.3 & 70.7 &    - &    - &    - &    - \\
    \mcl{Spindle\cite{zhao2017spindle}}     & CVPR17        &    - & 76.9 &    - &    - &    - &    - \\
    \mcl{Consistent Aware\cite{ConsAw}}     & CVPR17	    & 55.6 & 80.9 &    - &    - &    - &    - \\
    \mcl{GAN\cite{GAN}}                     & ICCV17    	& 56.2 & 78.1 & 47.1 & 67.7 &    - &    - \\
    \mcl{Latent Parts \cite{Li_2017_CVPR}}  & CVPR17        & 57.5 & 80.3 &    - &    - & 56.1 & 71.8 \\
    \mcl{ResNet+OIM \cite{OIM}}	            & CVPR17 		&    - & 82.1 &    - & 68.1 &    - &    - \\
    \mcl{ACRN\cite{ACRN}}                   & CVPR17-W 		& 62.6 & 83.6 & 52.0 & 72.6 &    - &    - \\
    \mcl{SVD \cite{SVD}}                    & ICCV17  		& 62.1 & 82.3 & 56.8 & 76.7 &    - &    - \\
    \mcl{Part Aligned \cite{zhao2017deeply}}& ICCV17        & 63.4 & 81.0 &    - &    - &    - &    - \\
    \mcl{PDC \cite{su2017pose}} 	        & ICCV17		& 63.4 & 84.1 &    - &    - &    - &    - \\
    \mcl{JLML \cite{JLML}}                  & IJCAI17       & 65.5 & 85.1 &    - &    - &    - &    - \\  
    \mcl{DPFL \cite{DPFL}}                  & ICCV17-W  &\f{72.6}&\f{88.6}& 60.6 & 79.2 &    - &    - \\
    \mcl{Forest \cite{Forest}}              & CVPR17    	&    - &    - &    - &    - & 50.7 & 70.6 \\
    \mcl{DGM+IDE \cite{DGM}}	            & ICCV17 		&    - &    - &    - &    - & 46.8 & 65.2 \\
	\hline
    \multirow{2}{*}{\rotatebox{90}{\hspace{-.1cm}Our}}
    & \mcl{ResNet-50 Baseline}                              & 59.8 & 82.6 & 50.3 & 71.5 & 49.5 & 64.5 \\
    & \mcl{\textbf{PSE}}   	                                &69.0&87.7&\f{62.0}&\f{79.8}&\f{56.9}&\f{72.1}\\

    \hline\hline
    \mcl{Smoothed Manif. \cite{Bai_2017_CVPR}}          & CVPR17      & 68.8 & 82.2 &    - &    - &    - &    - \\
    \mcl{IDE (R)+XQDA+k-reciprocal \cite{zhong2017re}}  & CVPR17      & 61.9 & 75.1 &    - &    - & 68.5 & 73.9 \\ 
    \mcl{IDE (R)+KISSME+k-reciprocal \cite{zhong2017re}}& CVPR17      & 63.6 & 77.1 &    - &    - & 67.3 & 72.3 \\ 
    \mcl{DaF \cite{yu2017divide}}                       & BMVC17      & 72.4 & 82.3 &    - &    - &    - &    - \\
    \hline
    \multirow{4}{*}{\rotatebox{90}{\hspace{-.1cm}Our}}
    & \multicolumn{2}{|l|}{PSE+ k-reciprocal \cite{zhong2017re}}				          & 83.5 & 90.2 & 78.9 & 84.4 &    70.7 &	74.9   \\
    & \multicolumn{2}{|l|}{PSE+ rank-dist (Eq. \ref{eq:ecn4})} 		  & 80.5 & 89.6 & 74.5 & 82.8 &   67.7 &	74.9  \\
    & \multicolumn{2}{|l|}{PSE+ ECN (orig-dist)}	                      & 80.5&\f{90.4}& 75.7& 84.5 & 68.6 & 75.5 \\
    & \multicolumn{2}{|l|}{\textbf{PSE+ ECN (rank-dist)}}	          &\f{84.0}&90.3&\f{79.8}&\f{85.2} & \f{71.8} & \f{76.7} \\

	\hline
 
\end{tabular}
}
\vspace{.1cm}
\caption{Comparison of our approach with the published state-of-the-art. The top section of the table compares our embedding with state-of-the-art approaches that do not use re-ranking. The lower section compares our combination of embedding and re-ranking to other state-of-the-art methods that use re-ranking.}
\label{tab:SOTA}
\vspace{-.1cm}
\end{table*}


\noindent{\textbf{Real World Considerations:}} In real-world applications re-id methods needs to be scalable (large gallery sizes) and are used in combination with automatic person detectors which can generate errors, such as mis-aligned detections or false positives. To investigate the scalability of our proposed PSE model, we evaluate on the Market+500k dataset to judge its robustness in real world deployment with very large galleries. The Market+500k dataset extends the Market dataset by including up to 500,000 distractor persons images. The relative change in mAP and rank-1 accuracy of our PSE model in comparison to other state-of-the-art approaches is depicted in Table \ref{tab:500k}.
While our embedding outperforms the published state-of-the-art without any distractors, the drop in accuracy observed when adding distractors is also notably less steep than that of other approaches. At 500,000 distractors our PSE's mAP has dropped by $12.5$\% while related approaches dropped by more than 14\%, similarly PSE drops in rank-1 accuracy by ${\sim}7$\% while the related approaches drop by ${\sim}10$\%. This shows the quality of our PSE model for this more realistic setting.

\begin{table}
\centering
\resizebox{\columnwidth}{!}{ 
\begin{tabular}{ | l|l  | cccc | cccc |}
    \hline
    \mcl{Method} & \multicolumn{4}{|c|}{mAP by \#Distractors} & \multicolumn{4}{|c|}{R-1 by \#Distractors} \\

    \mcl{}                          &    0 & 100k & 200k & 500k &    0 & 100k & 200k & 500k \\
	\hline\hline
    \mcl{I+V\np \cite{IV}} 			& 59.9 & 52.3 & 49.1 & 45.2 & 79.5 & 73.8 & 71.5 & 68.3 \\
    \mcl{APR\np$^*$ \cite{attid}} 	& 62.8 & 56.5 & 53.6 & 49.8 & 84.0 & 79.9 & 78.2 & 75.4 \\
    \mcl{TriNet\np$^\S$ \cite{trinet}} 	&\f{69.1}&61.9& 58.7 & 53.6 & 84.9 & 79.7 & 77.9 & 74.7 \\
    \hline
    \multirow{4}{*}{\rotatebox{90}{\hspace{-.1cm}Our}}
    & ResNet-50 Baseline 	        & 59.8 & 54.6 & 51.8 & 47.5 & 82.6 & 77.7 & 75.7 & 73.2 \\
    & Views Only 			        & 66.9 & 61.5 & 58.9 & 54.8 & \f{88.2} & \f{84.4} & \f{83.2} & \f{81.2} \\
    & Pose Only 			        & 63.0 & 57.7 & 54.9 & 50.6 & 83.6 & 80.0 & 77.9 & 75.1 \\
    & PSE 			                & 69.0 & \f{63.4} & \f{60.8} & \f{56.5} & 87.7 & 84.1 & 82.6 & 80.8 \\
	\hline
\end{tabular}
}
\vspace{.1cm}
\caption{Results of the PSE embedding on the Market-1501+500k distractors dataset ($\dag$ = unpublished works, $*$ = additional attribute ground truth, $\S$ = x10 test-time augmentation).}
\label{tab:500k}
\vspace{-.2cm}
\end{table}

	\begin{table}
    	\centering
        \resizebox{\linewidth}{!}{
        \begin{tabular}{| l | l | ccc | ccc | ccc | ccc |}
            \hline
            Detector 	& Method 		& \multicolumn{3}{c|}{\#detections=3} 	& \multicolumn{3}{c|}{\#detections=5} 	& \multicolumn{3}{c|}{\#detections=10}  \\
                        & & mAP & R-1 & R-20 						& mAP & R-1 & R-20 						& mAP & R-1 & R-20 						 \\
            \hline\hline
            DPM & IDE$_{det}$ \cite{PRW}			& 17.2 		& 45.9 		& 77.9 		& 18.8  	& 45.9 		& 77.4 		& 19.2  	& 45.7 		& 76.0		 \\
            DPM-Alex & IDE$_{det}$ \cite{PRW} 		& 20.2 		& 48.2 		& 78.1 		& 20.3 		& 47.4		& 77.1		& 19.9 		& 47.2 		& 76.4		 \\
            DPM-Alex & IDE$_{det}$+CWS \cite{PRW} 	& 20.0 		& 48.2 		& 78.8 		& 20.5 		& 48.3		& 78.8		& 20.5 		& 48.3 		& 78.8	\\
             \multicolumn{2}{|l|}{IAN (ResNet-101) \cite{IAN}} 	& 23.0 		& 61.9		&- 			&- 			&- 			&- 			&-			&- 			&- \\
            \hline
            DPM & Baseline 					& 25.4		& 59.0		& 83.9		& 27.5		& 59.1		& 83.9		& 28.3		& 58.1		& 83.3		\\
            DPM & View only 				& 28.5		& 63.4		& 87.3		& 30.8		& 63.1		& 86.8		& 31.4		& 62.0		& 86.1		\\
            DPM & Pose only 				& 26.2		& 59.1		& 84.6 		& 28.4		& 58.6		& 84.4		& 29.1		& 58.1		& 83.4	  	\\
            DPM & PSE						& \f{29.3}	& \f{65.1}	& \f{88.3}	& \f{31.7}	& \f{65.0}	& \f{88.2}	& \f{32.4}	& \f{64.5}	& \f{87.5}	\\
            \hline
        \end{tabular}
        }
        \vspace{.1cm}
        \caption{Results of PSE on PRW dataset (robustness against false detections): Considering 3, 5 and 10 detections per image.}
        \label{tab:prw}
        \vspace{-.3cm}
 \end{table}

In order to test our PSE embedding under detector errors, we train and evaluate its performance on the PRW dataset \cite{{PRW}}. 
Using the DPM detections provided with the dataset we observe similar trends as on Market or Duke.
Both types of pose information improve notably over the baseline and achieve a further increase in accuracy when combined in the PSE embedding. The performance is stable when considering more detections per image (hence increasing false positives) as shown in Table \ref{tab:prw}.
The PSE embedding achieves state-of-the-art accuracy, outperforming related approaches by at least 6.3\% in mAP (when an average of 3 detections per image are considered).
The results confirm the intuition that pose information is a helpful cue in identifying and handling mis-aligned and false-positive person detections.

%% file: sections/conclusion.tex
\section{Conclusion}
\label{sec:conclusion}
We have presented two related but independent contributions for person re-id and retrieval applications. 
We showed that both the fine and coarse body pose cues are important for re-id and proposed a new pose-sensitive CNN embedding which incorporates these. The PSE model currently relies on an external pose predictor, it would be useful to fully integrate this into the model.
The re-ranking method is unsupervised and can be used for general image and video retrieval applications. Both our person re-id model and re-ranking method set new state-of-the-art on a number of challenging datasets independently and in concert with each other. 


\noindent
\textbf{Acknowledgement:} The research described in this work is funded in part by the BMBF grant No. 13N14029. 